\documentclass[acmsmall,screen,nonacm]{acmart}
\usepackage{booktabs,tabularx,array,graphicx,float,amsmath,needspace,tikz,algorithm,algpseudocode}
\usepackage{placeins}
\usetikzlibrary{arrows.meta,positioning,calc,backgrounds,fit}
\newcolumntype{Y}{>{\raggedright\arraybackslash}X}
\newcolumntype{P}[1]{>{\raggedright\arraybackslash}p{#1}}

\newcommand{\flag}[1]{\texttt{#1}}

\newcommand{\tight}{\small\renewcommand{\arraystretch}{1.16}\setlength{\tabcolsep}{4pt}}
\makeatletter
\renewcommand{\@parfont}{\normalfont\bfseries}
\let\paper@startsection\@startsection
\def\@startsection#1#2#3#4#5#6{%
  \ifstrequal{#1}{paragraph}
    {\paper@startsection{#1}{#2}{#3}{\z@}{#5}{#6}}%
    {\paper@startsection{#1}{#2}{#3}{#4}{#5}{#6}}}
\makeatother
\begin{document}
\title{Agent-Guided Configuration Tuning for SMT Solvers}
\author{Cong Wang}
\email{wcon006@gmail.com}
\affiliation{\institution{Tsinghua University}\country{China}}
\author{Qiaoyi Zheng}
\email{zhengqiaoyi@sjtu.edu.cn}
\affiliation{\institution{Shanghai Jiao Tong University}\country{China}}
\author{Zhaoguo Wang}
\email{zhaoguowang@sjtu.edu.cn}
\affiliation{\institution{Shanghai Jiao Tong University}\country{China}}
\author{Fan Long}
\email{fanl@cs.toronto.edu}
\affiliation{\institution{University of Toronto}\country{Canada}}
\begin{abstract}
SMT solvers expose many configuration choices that determine how well their existing algorithms perform on a workload. We present, to our knowledge, the first workflow that uses general-purpose LLM agents to tune SMT solver configurations. Agents read solver documentation, source code, and persistent experiment records to propose option changes, and a deterministic harness measures each proposal, validates the returned models, and selects changes separately for each competition Division. Over roughly nine weeks, our tuning campaign recorded 4,545 proposals for cvc5 across 13 Divisions and 36 logics.

We evaluate the tuned configurations on the 2026 and 2025 SMT-COMP ModelValidation benchmarks under a 24-second cutoff. On problems not used for training, they validate more problems than official cvc5 in ten of the 13 Divisions in 2026 and in eight in 2025, with geometric-mean gains of 7.8\% and 6.8\% across Divisions. On the complete official selections, which include training problems, the tuned floating-point counts exceed those of every individual competition participant in both years. Further studies show how configuration changes make existing algorithms effective by changing access to strategies, scheduling, and option compatibility, and how benchmark families trade off against one another. The agents' experiment traces contain hypothesis-based proposals in every tuning group, and 64.3\% of the extracted hypotheses draw on prior measurements. These results show that agent-guided tuning can specialize a mature SMT solver and that accumulated experimental evidence informs sustained search.
\end{abstract}
\maketitle
\raggedbottom
\setlength{\emergencystretch}{1em}

\section{Introduction}
Satisfiability modulo theories (SMT) solvers decide whether logical formulas over theories such as arithmetic, bit-vectors, and arrays are satisfiable. They support program verification, symbolic execution, and test generation, where solver performance can determine the scale of the analysis. To handle different classes of constraints, they combine decision procedures, preprocessing passes, and search strategies~\cite{barbosa2022cvc5}. Their configuration options control how these components interact: which transformations to apply, which algorithms to enable, and how to divide execution time among strategies. These choices affect how well a solver's existing algorithms perform on a particular workload.

Configuration tuning can substantially improve solver performance on particular benchmark families. Algorithm configuration methods search for parameter settings using measured performance on a target set of problems~\cite{hutter2009paramils,hutter2011smac}. For example, automatic tuning of the SPEAR solver improved its performance on bounded model-checking and software-verification problems compared with extensively hand-tuned settings~\cite{hutter2007spear}. The same study found that effective settings depend on the benchmark family and on interactions among the solver's internal mechanisms. A setting that works on one workload may fail on another.

Configuration choices interact in ways that complicate tuning. An option may affect only one strategy, depend on another option, or prevent a strategy from running with model generation enabled. A portfolio that runs several solver strategies can contain an effective algorithm but start it too late to matter under an external time limit. A change that helps one benchmark family can hurt another. Source inspection and prior measurements can suggest useful changes, but evaluating a change requires checking both its aggregate effect on the workload and the validity of the solver's outputs.

\paragraph{\normalfont\bfseries Agent-guided solver specialization.}
We present, to our knowledge, the first workflow that uses general-purpose LLM agents to tune SMT solver configurations. Such agents can inspect a solver's configuration options alongside its source code and experiment history~\cite{yang2024sweagent}. They can connect information about a workload and the solver's implementation to configuration changes that can be measured directly. We apply this workflow to cvc5~\cite{barbosa2022cvc5} across the 13 Divisions of the SMT competition's ModelValidation track~\cite{smtcomp2026}, in which a \texttt{sat} answer must be accompanied by a model that can be validated. These Divisions cover arithmetic, bit-vectors, floating point, datatypes, and combinations with equality. Agents read solver documentation, source code, and persistent experiment records, then propose bounded configuration edits for their assigned Divisions. In the final workflow, a deterministic harness measures each proposal on its complete training group, validates models, and decides which changes to retain. Over roughly nine weeks, our tuning campaign recorded 4,545 proposals across 13 Divisions and 36 logics. The tuned configurations improve substantially on official cvc5 in several comparisons. For nonlinear integer arithmetic (NIA), on the 2026 test set of official problems not used for training, the tuned configuration validates 640 problems within 24 seconds, compared with 481 official cvc5 answers within the same cutoff, a 33.1\% increase.

Over thousands of rounds, the workflow must let agents pursue varied, informed proposals while preserving a consistent basis for evaluation and acceptance. Fresh agent sessions need access to earlier changes and measurements. A proposal can contain changes for several Divisions, although only some of those changes may be worth retaining. Apparent gains can also arise from measurement variation, and a higher answer count is insufficient when the outputs cannot be validated. A sustained campaign therefore requires explicit rules for evidence, for which Divisions each agent may edit, for correctness, and for the state carried from one round to the next.

The final workflow lets agents build on persistent experiment records while keeping evaluation and acceptance under fixed rules. Agents propose configuration edits for their assigned Divisions; the harness measures each proposal on the complete training group and validates the resulting models~\cite{beyer2019benchexec,bury2021dolmen}. Changes are selected separately for each Division based on model validity and measured performance. Preserved configurations, results, and decisions carry evidence between sessions and support recovery after interruptions. These mechanisms organize sustained search without eliminating measurement noise or changing the conventional propose-and-evaluate pattern~\cite{hutter2009paramils,hutter2011smac}.

\paragraph{\normalfont\bfseries Experimental results.}
We evaluate the tuned configurations on the benchmarks used in the 2026 and 2025 SMT-COMP ModelValidation tracks~\cite{smtcomp2026,smtcomp2025}. On problems not used for training, we compute, for each Division, the ratio of the tuned validated count to the official cvc5 \texttt{sat}-answer count. Under a 24-second cutoff, the geometric mean of these ratios improves by 7.8\% in 2026 and 6.8\% in 2025. Counts improve in ten of the thirteen Divisions in 2026 and eight in 2025. On the complete official selections, the tuned configurations also exceed the best competition participant's 24-second count in floating point in both years (6,234 versus 6,220 problems in 2026; 24,332 versus 24,321 in 2025).\footnote{The geometric mean weights all thirteen Divisions equally. The complete selections used for the all-participant comparison include training problems.}

Additional studies show that configuration changes make existing algorithms effective by changing when and how they are used. Source analysis and controlled experiments reveal that a solver can contain useful strategies yet fail to execute them within the available time or with the required output mode. Configuration tuning can therefore improve access to existing capabilities, and the mix of benchmark families in the workload determines which combinations are beneficial (Section~\ref{sec:eval-mechanisms}). We also analyze the agents' experiment traces; they contain hypothesis-based proposals in every tuning group. Of the extracted hypotheses, 64.3\% draw on prior measurements, and source code and documentation provide additional explanations of how options may affect solver behavior. The traces show agents connecting accumulated evidence to hypotheses about later changes, so part of the search is hypothesis-based (Section~\ref{sec:eval-hypotheses}).

\Needspace{8\baselineskip}
\paragraph{\normalfont\bfseries Contributions.}
This paper makes three contributions:
\begin{itemize}
  \item \textbf{The first LLM-agent workflow for SMT solver configuration tuning.} To our knowledge, this is the first workflow to apply general-purpose LLM agents to tuning SMT solver configurations. It combines the agents' access to documentation, source code, and experiment history with bounded edits, complete-group measurement, model validation, per-Division selection, and recovery for long-running search (Sections~\ref{sec:approach-proposal}--\ref{sec:approach-state}).
  \item \textbf{Empirical evidence of substantial specialization in a mature solver.} Evaluations across 13 Divisions and two years of official selections show improvements, including a 33.1\% increase on the 2026 NIA test set. We state training overlap and evaluation conditions explicitly (Section~\ref{sec:eval-outcomes}).
  \item \textbf{An explanation of solver behavior and agent search.} Source analysis, controlled experiments, family-level measurements, and trace analysis show how scheduling and option interactions affect performance. They also show how agents formulate hypotheses and how measurement and acceptance rules shape sustained search (Sections~\ref{sec:eval-mechanisms}--\ref{sec:eval-diagnostics}).
\end{itemize}

\section{Overview}
\label{sec:overview}

cvc5 is an open-source SMT solver developed from the CVC4 code base, with support for a broad range of theories and applications~\cite{barbosa2022cvc5}. Given a formula, it attempts to find an assignment satisfying its assertions or establish that none exists. For satisfiable problems, model generation returns an assignment that can be checked independently. Our goal is to specialize cvc5 through configuration tuning to maximize the number of problems with validated models under a fixed time budget. We use the cvc5 SMT-COMP 2025 submission, version \texttt{1.3.0-modified} (Git revision \texttt{b75ccda88}), as the target build and tune the options it exposes.

\paragraph{\normalfont\bfseries Configurable solver mechanisms.}
For the quantifier-free problems in this study, cvc5 combines preprocessing, Boolean search, and theory reasoning. Preprocessing simplifies assertions and can change their representation, for example by translating bit-vector constraints into Boolean constraints. Boolean search explores truth assignments to atoms; theory solvers check whether the assigned atoms are consistent and return lemmas that guide later search. Options control both the individual procedures and their interaction. A rewriting option can change the constraints that reach a theory solver; a decision heuristic can change which theory facts are considered first. This structure gives an agent several ways to specialize the solver's behavior without changing its implementation~\cite{barbosa2022cvc5}.

The target build also provides a portfolio of predefined strategies, chosen according to the problem's logic. Each strategy is a combination of option values, and the portfolio controls when strategies start and how long they run. A configuration can change the number of concurrent portfolio workers, or disable the portfolio and set options directly. An option edit may therefore alter access to an algorithm as well as how that algorithm behaves once it runs. The evaluation examines both effects.

\paragraph{\normalfont\bfseries Workload and configuration scope.}
We use SMT-COMP ModelValidation workloads~\cite{smtcomp2026}. Problems are written in SMT-LIB and declare a logic that identifies the permitted theories~\cite{barrett2016smtlib}. The competition groups related logics into Divisions. Our objective is to maximize the number of problems for which the solver returns a model within 24 seconds and the Dolmen validator accepts that model. The official ModelValidation results report performance at 24 seconds for runs that permit 1,200 seconds, and we target this short-budget setting. For comparison, we filter official records at 24 seconds; locally, we enforce a hard 24-second wall-clock limit. A \emph{solved} problem has a timely \texttt{sat} answer; a \emph{validated} problem also has a model accepted by Dolmen.

A configuration supplies shared options, Division options, and logic options, applied in that order, so later values override earlier ones. It may specialize behavior by logic but may not choose options based on a problem's identity, family, path, or expected answer. The harness owns resource limits and the solver protocol. We partition the Divisions into groups, each with its own configuration, training problems, and tuning loop. A group may contain several Divisions, but acceptance compares each Division with its own incumbent record. The initial configuration uses the target build's official ModelValidation options, with up to two additional options per logic. A single campaign spans the three phases in Table~\ref{tab:phases}: configurations carry forward, and each phase measures their scores under its own conditions.

After excluding harness-owned options, the target build exposes 494 distinct options. The tuned configurations change 78 distinct options across the studied logics, including options for preprocessing, arithmetic, bit-vectors, and portfolio scheduling. Not all exposed options are relevant to these workloads: the 346 options in the Other category largely concern quantifiers, strings, proofs, and input and output. Appendix~\ref{app:configuration-scope} in the supplementary material gives the category breakdown and counting conventions.

\paragraph{\normalfont\bfseries Workflow.}
We implement the tuning workflow with Humanize, a framework for running coding agents in iterative workflows~\cite{humanize}. Figure~\ref{fig:workflow}(b) shows one round of the final workflow within a tuning group. The harness records the group's current configuration, the \emph{incumbent}, as the round's \emph{parent}. In step 1, the active agent reads persistent state, documentation, and solver source, then proposes a complete \emph{candidate} configuration by editing the parent's options (\S\ref{sec:approach-proposal}). Before measurement, the harness checks the candidate; after a failed check, the same agent repairs it (\S\ref{sec:approach-checks}). In step 2, the harness measures the complete training group and validates models; in step 3, it keeps only the edits of the Divisions that satisfy the acceptance rule and restores the other changed sections (\S\ref{sec:approach-measurement}). In step 4, the harness saves the resulting incumbent and the round's evidence for the next round (\S\ref{sec:approach-state}). The same active agent handles proposal and repair; checks, evaluation, selection, and record updates are deterministic harness operations.

We organize Divisions into tuning groups to simplify resource allocation and roughly balance progress across workloads of different sizes. Divisions contain different numbers of training problems, so allocating identical resources to independent per-Division loops could let smaller workloads advance through many more rounds than larger ones. Grouping smaller workloads and allocating resources at the group level lets us account for these differences without continually rescheduling individual Divisions. A shared agent session can also reuse context when proposing changes for several Divisions, which may reduce token usage. These motivations concern deployment and do not guarantee equal tuning effort: the agent chooses which Divisions to edit in each round, and acceptance remains per Division.

\begin{figure}[htbp]
\centering
\definecolor{hzblue}{HTML}{2F6FD0}\definecolor{hzbluebg}{HTML}{EAF1FB}
\definecolor{hzorange}{HTML}{E0663C}\definecolor{hzorangebg}{HTML}{FCEEE7}
\definecolor{hzdoc}{HTML}{EFEFEB}\definecolor{hzband}{HTML}{E3E3DF}
\definecolor{hzred}{HTML}{D23B3B}\definecolor{hzloop}{HTML}{6A3D9A}\definecolor{hzgreen}{HTML}{2F9A45}
\def\wfdd{-{}-}
\def\wfdoc#1#2#3#4{%
  \filldraw[fill=hzdoc, draw=black!50, line width=0.5pt]
    (#1,#2) -- ({#3-0.15},#2) -- (#3,{#2-0.15}) -- (#3,#4) -- (#1,#4) -- cycle;
  \filldraw[fill=white, draw=black!50, line width=0.45pt]
    ({#3-0.15},#2) -- ({#3-0.15},{#2-0.15}) -- (#3,{#2-0.15}) -- cycle;}
\begin{tikzpicture}[
  x=1cm, y=1cm, font=\sffamily\footnotesize,
  >={Stealth[length=1.6mm,width=1.3mm]},
  flow/.style={->, line width=0.55pt, draw=black!80},
  back/.style={->, line width=0.55pt, draw=black!60, dash pattern=on 2.2pt off 1.6pt},
  mainc/.style={->, line width=0.9pt, draw=hzloop},
  actor/.style={rounded corners=2.5pt, line width=0.7pt},
  codex/.style={actor, draw=hzorange, fill=hzorangebg},
  claude/.style={actor, draw=hzblue, fill=hzbluebg},
  tool/.style={actor, draw=black!55, fill=white},
  hd/.style={font=\sffamily\footnotesize\bfseries, anchor=base west, inner sep=0pt},
  ttl/.style={font=\sffamily\footnotesize\bfseries, anchor=base west, inner sep=0pt},
  sub/.style={font=\sffamily\scriptsize, text=black!62, anchor=north west, inner sep=0pt, align=left,
              execute at begin node={\hyphenpenalty=10000\exhyphenpenalty=10000}},
  lbl/.style={font=\sffamily\scriptsize, text=black!68, inner sep=1.2pt},
  num/.style={circle, draw=black!80, line width=0.5pt, inner sep=0pt, minimum size=3.2mm,
              font=\sffamily\scriptsize\bfseries},
  cell/.style={font=\sffamily\scriptsize, anchor=west, inner sep=0pt},
  code/.style={font=\ttfamily\scriptsize, anchor=west, inner sep=0pt},
  exreg/.style={rounded corners=2.5pt, line width=0.6pt, draw=black!45, dash pattern=on 2pt off 1.5pt},
  yes/.pic={\draw[green!45!black, line width=0.8pt, line cap=round, line join=round]
              (-0.065,0.005) -- (-0.02,-0.05) -- (0.075,0.07);},
  no/.pic={\draw[red!65!black, line width=0.8pt, line cap=round]
              (-0.055,-0.055) -- (0.055,0.055) (-0.055,0.055) -- (0.055,-0.055);},
  gate/.pic={\fill[hzred] (22.5:0.2) \foreach \a in {67.5,112.5,...,337.5} {-- (\a:0.2)} -- cycle;
             \fill[white] (-0.11,-0.03) rectangle (0.11,0.03);},
  ok/.pic={\filldraw[fill=white, draw=hzgreen, line width=0.7pt] (0,0) circle (0.16);
           \draw[hzgreen, line width=0.9pt, line cap=round, line join=round]
             (-0.075,0.0) -- (-0.02,-0.06) -- (0.08,0.065);},
]

\node[hd] at (0,-0.4) {(a)\enspace\mdseries Flame Chase};
\wfdoc{2.45}{-0.1}{3.2}{-0.62}   \node at (2.8,-0.38) {$S_r$};
\draw[flow] (3.2,-0.36) -- (3.45,-0.36);
\draw[codex] (3.45,-0.1) rectangle (5.2,-0.62);
  \node[align=center, font=\sffamily\scriptsize, inner sep=0pt] at (4.325,-0.36) {\footnotesize Codex\\[-1pt]{\color{black!60}round $r$}};
\draw[flow] (5.2,-0.36) -- (5.45,-0.36);
\wfdoc{5.45}{-0.1}{6.2}{-0.62}   \node at (5.8,-0.38) {$S_{r+1}$};
\draw[flow] (6.2,-0.36) -- (6.45,-0.36);
\draw[claude] (6.45,-0.1) rectangle (8.35,-0.62);
  \node[align=center, font=\sffamily\scriptsize, inner sep=0pt] at (7.4,-0.36) {\footnotesize Claude Code\\[-1pt]{\color{black!60}round $r{+}1$}};
\draw[flow] (8.35,-0.36) -- (8.6,-0.36);
\wfdoc{8.6}{-0.1}{9.35}{-0.62}   \node at (8.95,-0.38) {$S_{r+2}$};
\draw[flow] (9.35,-0.36) -- (9.6,-0.36);
\draw[codex] (9.6,-0.1) rectangle (11.35,-0.62);
  \node[align=center, font=\sffamily\scriptsize, inner sep=0pt] at (10.475,-0.36) {\footnotesize Codex\\[-1pt]{\color{black!60}round $r{+}2$}};
\draw[flow] (11.35,-0.36) -- (11.6,-0.36);
\node[anchor=west, inner sep=0pt] at (11.7,-0.36) {$\cdots$};
\node[sub, text width=11.2cm] at (2.45,-0.78)
  {Each round starts a fresh session. The working directory $S$ (incumbent configuration, per-problem results, Division decisions, notes) is all that passes from one round to the next.};

\node[hd] at (0,-1.62) {(b)\enspace\mdseries One round};

\draw[actor, draw=black!70, fill=white] (0,-1.85) rectangle (1.75,-5.85);
\node[font=\sffamily\footnotesize\bfseries, inner sep=0pt] at (0.875,-3.1) {Agent};
\node[codex, inner xsep=4pt, inner ysep=1.5pt, font=\sffamily\scriptsize] at (0.875,-3.63) {Codex};
\node[lbl] at (0.875,-3.96) {or};
\node[claude, inner xsep=4pt, inner ysep=1.5pt, font=\sffamily\scriptsize] at (0.875,-4.29) {Claude Code};
\node[lbl, align=center] at (0.875,-4.8) {the agent of\\[-1pt]this round};

\wfdoc{1.95}{-2.3}{4.95}{-3.5}
\node[num] at (2.17,-2.52) {1};
\node[ttl] at (2.36,-2.58) {candidate};
\node[sub, text width=2.75cm] at (2.1,-2.75) {at most one argument section per Division};
\draw[mainc] (1.75,-2.9) -- (1.95,-2.9);
\draw[mainc] (4.95,-2.9) -- (5.55,-2.9);
\pic at (5.75,-2.9) {gate};
\node[lbl, inner sep=0pt] at (5.75,-3.27) {checks};
\draw[mainc] (5.95,-2.9) -- (6.5,-2.9);
\draw[back] (5.75,-2.7) -- (5.75,-2.0) -- (1.75,-2.0);
\node[lbl, anchor=south west, align=left, inner sep=1pt] at (1.95,-2.0) {failed check: the same agent\\[-1pt]repairs it in the same round};

\draw[actor, draw=black!60, fill=white] (6.5,-2.3) rectangle (13.6,-3.5);
\node[num] at (6.72,-2.52) {2};
\node[ttl] at (6.91,-2.58) {measure};
\node[sub, text width=6.75cm] at (6.65,-2.75)
  {\textbf{BenchExec} runs every training problem of the group, 24\,s wall and 96\,s CPU time each; \textbf{Dolmen} checks every \texttt{sat} model};

\draw[mainc] (12.75,-3.5) -- (12.75,-4.5);
\node[lbl, anchor=west] at (12.83,-4.0) {results};

\draw[actor, draw=black!60, fill=white] (6.5,-4.5) rectangle (13.6,-5.7);
\node[num] at (6.72,-4.72) {3};
\node[ttl] at (6.91,-4.78) {select per Division};
\node[sub, text width=6.75cm] at (6.65,-4.95)
  {update only the Divisions that are eligible and above their incumbent record; restore the other changed sections};

\wfdoc{1.95}{-4.5}{4.95}{-5.7}
\node[num] at (2.17,-4.72) {4};
\node[ttl] at (2.36,-4.78) {state $S_{r+1}$};
\node[sub, text width=2.75cm] at (2.1,-4.95) {new incumbent, results, and Division decisions};
\draw[mainc] (6.5,-5.1) -- node[above, font=\sffamily\scriptsize\bfseries, text=hzloop, inner sep=1.5pt] {feedback} (4.95,-5.1);
\draw[mainc] (1.95,-5.1) -- (1.75,-5.1);

\draw[hzloop, line width=1.6pt, -{Stealth[length=2.4mm,width=2.2mm]}]
  (5.75,-4.0) ++(70:0.33) arc[start angle=70, end angle=-245, radius=0.33];
\node[anchor=west, align=left, inner sep=0pt, font=\sffamily\scriptsize, text=hzloop] at (6.25,-4.0)
  {\textbf{\footnotesize tuning loop}\\[-0.5pt]one pass per round; the next round uses the other agent};

\begin{scope}[yshift=-6mm]
\draw[exreg] (0,-5.55) rectangle (13.8,-8.45);
\node[font=\sffamily\scriptsize, anchor=base west, inner sep=0pt] at (0.15,-5.82)
  {\textbf{\footnotesize (c)}\enspace\textbf{\footnotesize Example}\enspace recorded decisions of round r000144 (phase III, small-mix group), which edited five Divisions};

\foreach \x/\h in {0.2/Division, 3.75/Section, 5.1/Edit, 8.4/{Validated (candidate vs.\ incumbent)}, 12.6/Result}
  \node[cell, font=\sffamily\scriptsize\bfseries] at (\x,-6.14) {\h};
\draw[black!45, line width=0.4pt] (0.2,-6.3) -- (13.6,-6.3);
\draw[black!45, line width=0.4pt] (0.2,-8.25) -- (13.6,-8.25);

\foreach \y/\d/\s/\e/\m/\r in {
   -6.52/{Equality + bit-vectors}/{QF\_UFBV}/{\wfdd learned-rewrite}/{282 vs.\ 280}/1,
   -6.9/{Nonlinear real arithmetic}/{QF\_NRA}/{\wfdd repeat-simp}/{229 vs.\ 228}/1,
   -7.28/{ADT + linear arithmetic}/{QF\_ALIA}/{\wfdd learned-rewrite}/{258 vs.\ 259}/0,
   -7.66/{Equality + linear arithmetic}/{QF\_UFIDL}/{\wfdd decision=justification}/{281 vs.\ 300}/0,
   -8.04/{Linear real arithmetic}/{QF\_LRA}/{\wfdd decision=stoponly}/{251 vs.\ 251}/0} {
  \node[cell] at (0.2,\y) {\d};
  \node[code] at (3.75,\y) {\s};
  \node[code] at (5.1,\y) {\textsf{+}\,\e};
  \ifnum\r=1
    \pic at (8.48,\y) {yes}; \node[cell] at (8.63,\y) {\m};
    \node[cell, font=\sffamily\scriptsize\bfseries, text=green!35!black] at (12.6,\y) {updated};
  \else
    \pic at (8.48,\y) {no}; \node[cell] at (8.63,\y) {\m};
    \node[cell, text=black!65] at (12.6,\y) {restored};
  \fi
}
\end{scope}
\end{tikzpicture}
\caption{The final workflow. (a)~Flame Chase: Codex and Claude Code alternate rounds; each round starts a fresh session, and the working directory $S$ is all that passes from one round to the next. (b)~One round: the agent proposes a candidate (step 1); the harness checks it and returns a failed candidate to the same agent; the harness then measures the candidate on the complete group, validates the models (step 2), and updates only the Divisions that satisfy the acceptance rule (step 3); the new incumbent, results, and Division decisions form the state $S_{r+1}$ (step 4), and the tuning loop repeats with the other agent. (c)~The recorded decisions of phase-III round r000144, which edited five Divisions; each row compares the candidate's validated count with the Division's incumbent record. The two Divisions that improved were updated, and the other three sections were restored.}
\Description{Three panels. (a) A chain of rounds in which Codex and Claude Code alternate and only the working directory S passes between rounds. (b) One round: the agent's candidate passes a blocking check in the harness; BenchExec runs it on every training problem of the group and Dolmen checks the models; the harness then updates only the Divisions that satisfy the acceptance rule and saves the state for the next round. A candidate that fails a check returns to the same agent. The steps form one pass of the tuning loop; the next round uses the other agent. (c) A table of the decisions recorded for round r000144, which edited five Divisions, updated two of them, and restored the other three.}
\label{fig:workflow}
\end{figure}

\paragraph{\normalfont\bfseries One campaign, three phases.}
The workflow evolved through three phases (Table~\ref{tab:phases}). Phase I established the initial tuning loop. Starting in phase II, we enabled Humanize's Flame Chase flow~\cite{humanize-flame-chase}, in which two agents alternate rounds on the same task. Each round starts a fresh agent session, and progress carries forward through the shared working directory and its saved experiment records. In our deployment, the two agents use different language models. Phase II also introduced model validation and per-Division selection. Differences between the 2025 and 2026 workloads then led us to expand the training set, and phase III continued from the phase-II configurations. Figure~\ref{fig:workflow} shows the final workflow; Section~\ref{sec:approach} describes when its checks were introduced.

According to the available session metadata, the agents used \texttt{gpt-6-astra} and \texttt{claude-fable-5-1} in the early phases, and \texttt{gpt-6.1-sol} and \texttt{claude-opus-5-5} in phase III. Deployment records place the model-pair transition late in phase II. Many records lack a model field, so these identifiers describe the recorded deployment but do not give the model used in every round.

Phases I and II used 10,104 sampled 2025 problems that timed out at least once during a preliminary tuning run. This criterion selects for observed difficulty; the phase-I baseline still solved 892 of them. Phase III added 5,148 sampled 2026 problems, for 15,252 in total. Within each Division, the problems not already selected were ordered by the SHA-256 hash of the fixed integer seed 20261011, the Division, and the problem path; $n$ denotes the number of problems in this order. The first $\lfloor0.3n\rfloor$ were then taken without consulting run outcomes. All selected problems become training problems and are excluded from the test sets.

In phase III, floating point moves from a group shared with other Divisions to its own group. No phase-II score is carried into phase III. The 14 non-proposal records comprise six in phase I and eight in phase III, including six phase-III baselines and two operator fixes.

\begin{table}[htbp]
\caption{Chronological phases of the campaign. Records include proposals, baselines, and operator fixes; some proposals have no completed measurement.}
\label{tab:phases}
\tight
\begin{tabularx}{\linewidth}{@{}lP{.15\linewidth}Yrr@{}}
\toprule
Phase & Duration & Workload and transition & Records & Proposals \\
\midrule
I & About 2 weeks & 10,104 problems; five groups; host A; answer-based search & 517 & 511 \\
II & About 3 weeks & Same problems and host; inherited models validated; per-Division selection & 997 & 997 \\
III & About 3.5 weeks & 15,252 problems; six groups; phase-II configurations remeasured on hosts B/C & 3,045 & 3,037 \\
\midrule
Total & & 13 Divisions, 36 logics & 4,559 & 4,545 \\
\bottomrule
\end{tabularx}
\end{table}

\section{Approach}
\label{sec:approach}

Agents receive the information they need to propose informed changes, and the harness controls the measurement objective and state transitions. The main design decisions concern how fresh sessions use accumulated evidence, how each candidate receives a comparable measurement, and how independently accepted Division edits form the next incumbent. The subsections follow the order of Figure~\ref{fig:workflow}.

\paragraph{\normalfont\bfseries When safeguards appeared.}
Phase I initially counted non-timeout terminations, including crashes, and did not validate models or reliably restore rejected edits. In the second week of tuning, an operator audit changed the score to count actual answers. Phase II introduced Dolmen validation, per-Division eligibility and acceptance, deterministic restoration of rejected edits, and protected results with explicit recovery. From phase II on, the task rules prohibit direct solver probes and subset runs; phase I permitted both. At the transition, existing results were reclassified under validation, but a complete validated baseline from a new measurement is unavailable. Phase III remeasured its inherited configurations under the expanded workload and new hosts. Smoke checks were added near the start of phase III. The subsections below describe the round in Figure~\ref{fig:workflow}(b), including the candidate check (edit scope, option help, and smoke check) that runs between steps 1 and 2.

\subsection{Propose Edits}
\label{sec:approach-proposal}

\paragraph{\normalfont\bfseries Reasoning from persistent evidence.}
At the start of a round, the harness copies the incumbent into a new round directory and records that configuration as the parent. Under Flame Chase, the two agents alternate rounds within each group, and each round starts a fresh session with no conversation history. Within a round, one agent handles the proposal and any repairs after failed checks. The persistent state therefore has two roles. It identifies the configuration and measurements against which the proposal will be judged, and it carries information from earlier experiments into the next session. An agent can read configuration differences, per-problem outcomes, validation results, and acceptance decisions, together with notes left by earlier sessions. Rejected candidates remain part of this history because their family-level gains and losses may explain an option's behavior even when their total score did not improve.

The task directs agents to consult this evidence together with the official solver documentation, the target binary's option help, and the solver source code. Each source supports a different part of a proposal. Documentation states an option's intended effect, source inspection can reveal a dependency or an inactive code path, and prior measurements show where a proposed mechanism might matter. For example, a family-specific regression can motivate examining a theory option rather than changing unrelated search heuristics. The workflow does not prescribe an order in which to try options or require a particular search strategy. It asks for small, justified changes but does not mechanically require a formal hypothesis; the trace analysis measures the hypotheses that agents record.

\paragraph{\normalfont\bfseries A bounded configuration interface.}
An agent may edit only the argument sections of its assigned Divisions. In one round, it may change at most one argument section per Division: either the Division section or one of that Division's logic sections. A section can contain several option changes, and one proposal can affect several Divisions. Shared options, metadata, and the set and order of sections stay fixed. This constraint makes explicit the unit that the harness retains or restores.

Agents may not change the harness, validator, training selection, historical records, or resource limits. They may not inspect the test sets, run direct solver probes or selected benchmark subsets, or start extra benchmark processes. They may read the binary's option help. Under these rules, an agent can use earlier experiments to choose the next candidate, while new performance evidence comes from the same complete-group measurement. This separation fixes the meaning of the score and lets the proposal strategy vary with the evidence.

\subsection{Check Candidate}
\label{sec:approach-checks}

\paragraph{\normalfont\bfseries Checking executable proposals.}
Before complete-group measurement, the harness checks the edit scope and verifies each option against the target binary's option help. It rejects harness-owned options, and the agent task also prohibits settings with known failures. A smoke check then runs the two smallest training problems of each logic whose effective options changed, for up to 20 seconds each. Real problems exercise combinations that an option-name check alone cannot reject, such as an unavailable SAT back end or a setting incompatible with model output. An edit-scope violation or a detected option error sends the candidate back to the same active agent for repair, and the repaired candidate is checked again. A smoke-check timeout is inconclusive and does not block the candidate. These checks show that the candidate is suitable for measurement, not that its models will be valid.

\subsection{Evaluate and Select}
\label{sec:approach-measurement}
\label{sec:approach-acceptance}

In steps 2 and 3, the harness measures the complete group, validates models, and selects edits per Division. A poor result is handled by the acceptance rule below; the candidate does not return to the agent for repair. Failure or interruption of the overall measurement invokes recovery (\S\ref{sec:approach-state}).

\paragraph{\normalfont\bfseries Measuring validated answers.}
BenchExec~\cite{beyer2019benchexec} runs the candidate on every training problem in its group, without early stopping. A slot executes one problem at a time on four logical CPUs, with limits of 24 seconds wall time and 96 seconds CPU time. The allocation is fixed for the group. Dolmen~\cite{bury2021dolmen} checks the model of every \texttt{sat} answer, using the competition tools' validation procedure. Only timely answers with accepted models contribute to the validated count. Timeouts and \texttt{unknown} answers count as unsolved; under the ModelValidation workload rules, \texttt{unsat} counts as an error.

For a Division, \emph{eligibility} requires a complete measurement, no wrong answer or model, and no unverified \texttt{sat} answer. An unverified model differs from a model shown to be wrong, but either one makes the Division ineligible. An individual solver execution failure contributes no validated answer and does not by itself establish a wrong model. Before using a measurement's results, the harness also checks that the measurement run as a whole completed successfully. Validation, and therefore acceptance, covers only the measured training problems.

\paragraph{\normalfont\bfseries Comparing against the incumbent.}
Let $B_d$ be the pre-round incumbent record for Division $d$, including its eligibility and validated count. For a completed measurement $M$, the harness retains a change to $d$ only when
\begin{equation}
\label{eq:retain}
\operatorname{retain}(M,d,B_d)
= \operatorname{eligible}(M,d)\ \land\
\bigl(\neg\operatorname{eligible}(B_d)
\ \lor\ \operatorname{validated}(M,d)>\operatorname{validated}(B_d)\bigr).
\end{equation}
Replacing an eligible incumbent therefore requires a strict increase; on a tie, the Division keeps its parent configuration. An ineligible incumbent is a starting point for the search: the first eligible result can replace it even with a lower count. This rule gives validity priority over performance. Because the rule applies to each Division separately, a large gain in one Division cannot mask a regression in another Division that has an eligible incumbent, or make an invalid result acceptable.

\paragraph{\normalfont\bfseries Updating the retained Divisions.}
A proposal can change several Divisions, and the harness decides each one separately. It keeps only the edits of the Divisions that satisfy Equation~\eqref{eq:retain} and restores every other changed section to its parent value, so one round can update some Divisions and leave the others unchanged. Figure~\ref{fig:workflow}(c) shows a round that updated two of the five Divisions it edited.

The harness keeps separate records for Division acceptance and for the complete group configuration, so configurations, measurements, and decision thresholds stay distinct. A Division's incumbent record identifies the measurements that support its threshold. The group record describes one complete measurement of one configuration. A score increase in an unchanged Division does not create a new incumbent record, but evidence that an unchanged Division's incumbent is ineligible invalidates its record.\footnote{This rule covers proposals that change at least one Division. A separate rule updates the records when a byte-identical remeasurement of the entire incumbent reduces the number of unsolved problems.}

Execution on shared hosts varies, so requiring a strict improvement in one measurement can select a noisy gain. The evaluation therefore analyzes the recorded acceptance margins and later measurements separately from the acceptance rule used during tuning.

\subsection{Save Results}
\label{sec:approach-state}

After selection, the harness publishes the retained configuration as the next incumbent. It records the proposal, check results, raw benchmark outputs, per-problem validation results, Division decisions, and the agent trace. The next session can link an observed score to the configuration that produced it and to the resulting decision. Our analyses of solver behavior and of the agents' hypotheses therefore use these records directly.

The measurement command refuses to overwrite a directory that contains raw results. For an interrupted round, recovery first checks that the loop is no longer active and that the configuration hashes match the recorded state. It then preserves the partial results, restores the incumbent, and leaves the round unscored. A requested pause takes effect at a round boundary. Human operators maintain the deployment and authorize repairs; an agent's diagnosis does not authorize changes to the harness or its acceptance rule. These mechanisms keep the experiment history usable across the sessions of a long campaign without treating an incomplete run as a completed comparison.

\section{Evaluation}
\label{sec:evaluation}
The evaluation addresses three questions: how much the tuned configurations improve performance across SMT workloads (Section~\ref{sec:eval-outcomes}); which solver mechanisms explain why configuration changes can help (Section~\ref{sec:eval-mechanisms}); and how agents use experimental evidence to formulate and assess proposals during sustained search (Sections~\ref{sec:eval-progress}, \ref{sec:eval-hypotheses}, and~\ref{sec:eval-diagnostics}). We first describe the benchmarks and the measurement procedure.

\subsection{Study Scope and Resources}
\label{sec:eval-setup}
\paragraph{Campaign and configurations.}
The campaign spans roughly nine weeks and all three phases in Table~\ref{tab:phases}. We compare the tuned configurations with official cvc5. The configuration differences between the two include both the initial logic-specific choices and the changes selected during tuning; the other differences are build, host, and run mode.

The tuned configurations are those frozen for the November 6--7 evaluation, with three options removed for correctness: \flag{--early-ite-removal} on QF\_BV, \flag{--nl-icp} on the equality with nonlinear arithmetic Division, and \flag{--nl-ext-ent-conf} on QF\_UFNIA and QF\_UFDTNIA. The frozen configurations do not include two changes accepted later: \flag{--no-static-learning} for bit-vectors and \flag{--sat-solver=cadical} for equality. We evaluate on the complete official selections, using each year's benchmark files as scrambled for the competition. Results without the training problems are reported separately.

\paragraph{Benchmarks.}
We use the SMT-LIB benchmarks of the 2025 and 2026 SMT-COMP ModelValidation tracks~\cite{smtcomp2025,smtcomp2026} and draw our own sample of training problems from them. Phases I and II use 10,104 problems from the 2025 benchmarks that timed out at least once during a preliminary tuning run. Phase III adds 5,148 problems from the 2026 benchmarks, sampled at 30\% per Division using a fixed hash ordering, for 15,252 training problems. The training set concentrates on difficult problems, and phase III broadens its coverage.

For evaluation, we include every problem in the competition's official benchmark selections: 19,094 problems in 2026 and 59,750 in 2025. We report both these complete selections and their subsets without training problems. These subsets contain 12,498 and 51,671 problems, respectively; we call them the test sets. The complete selections measure performance on the competition workloads, and the test sets measure performance beyond the training problems. Training problems make up 6,596 of the 19,094 official 2026 problems and 8,079 of the 59,750 official 2025 problems. We exclude training problems by problem identity and have not established that test and training problems are disjoint in content. We keep the complete selections and the test sets separate throughout the comparisons.

\paragraph{Measurement and baselines.}
The target build is the cvc5 team's SMT-COMP 2025 submission, version \texttt{1.3.0-modified}. Each local problem run receives four logical CPUs, a 24-second wall-clock limit, and a 96-second CPU-time limit. BenchExec records execution results, and Dolmen checks the returned models. Our performance measure is the validated count: the number of problems for which the solver returns a model within the time limit and Dolmen accepts that model. Timeouts, unknown answers, and unverified models do not contribute. The official baseline is cvc5's \texttt{sat}-answer count in the official records, filtered at 24 seconds. Official runs allow up to 1,200 seconds, whereas local runs enforce the shorter limit directly. Each comparison therefore pairs a local count with a count from the official records; we did not rerun participants on a common host.

To summarize performance across Divisions, we compute, for each Division, the ratio of the tuned validated count to the official cvc5 \texttt{sat}-answer count, and take the geometric mean of the thirteen ratios. Each Division has equal weight, so the large floating-point Division does not dominate the aggregate. We also report per-Division counts and pooled totals; the pooled totals show how many additional problems are validated.

\paragraph{Execution resources.}
Table~\ref{tab:campaign} summarizes the campaign's coverage and measurement cost. Phase III uses 22 slots on two shared hosts, with four logical CPUs per slot. The recorded cost is about 24,220 slot hours, covering preparation, execution, collection, scoring, and validation. Slot hours measure allocated time, not CPU utilization. Campaign coverage extends to the end of tuning, whereas the timing and late-search analyses use the accounting snapshot from the day before tuning ended. Because of this difference in coverage, there are fewer timed proposal measurements than proposals.

Of the timed executions, 4,387 finished normally, 19 were marked bench-failed, and one aborted because the disk was full. Another 34 measurement records lack usable timing intervals; we do not impute them. Hosts were shared. Phases I--II enforced a memory cap of 30,720 MB, and phase III enforced none.

\begin{table}[htbp]
\caption{Coverage and resources. Campaign coverage reaches the end of tuning; cost figures use the accounting snapshot from the day before tuning ended and include failed executions with recorded timing intervals.}
\label{tab:campaign}
\tight
\begin{tabularx}{\linewidth}{@{}P{.36\linewidth}Y@{}}
\toprule
Quantity & Recorded value \\
\midrule
Campaign coverage & 4,559 records; 4,545 proposals; 13 Divisions and 36 logics \\
Training problems & Phases I--II: 10,104; phase III: 15,252 \\
Timed measurements & 4,407 execution intervals: 4,396 proposal, 11 baseline \\
Recorded measurement cost & 24,219.5 slot hours (about 24,220); four logical CPUs per slot \\
Phase-III allocation & 22 slots on two hosts; six groups; six slots for NIA \\
Per-problem limits & 24 s wall time and 96 s CPU time \\
\bottomrule
\end{tabularx}
\end{table}

\subsection{Outcomes across Divisions}
\label{sec:eval-outcomes}
We assess whether the tuned configurations improve cvc5 across Divisions beyond the training problems. Table~\ref{tab:test} compares the tuned configurations with official cvc5 on the two test sets, and Table~\ref{tab:full} gives the corresponding results on the complete selections.

\paragraph{Performance beyond the training set.}
The tuned configurations achieve geometric-mean increases of 7.8\% in 2026 and 6.8\% in 2025. Improvements occur in ten of the thirteen Divisions in 2026, with two ties and one regression, and in eight in 2025, with three ties and two regressions. In pooled totals, the tuned configurations validate 11,304 problems in 2026 and 47,394 in 2025, against 10,941 and 46,349 for official cvc5. The gains extend beyond the training problems and appear in both years.

NIA has the largest absolute test gains: 640 versus 481 problems in 2026 and 3,313 versus 2,758 in 2025, increases of 33.1\% and 20.1\%. Benefits also appear in different theory combinations. Equality with bit-vectors rises from 133 to 226 and from 111 to 190, and linear integer arithmetic (LIA) rises from 754 to 826 and from 3,124 to 3,477. Useful specialization spans several arithmetic procedures and benchmark families.

The improvements are not uniform. Nonlinear real arithmetic loses 52 test problems in 2026 and 229 in 2025, and the bit-vectors Division loses 84 in 2025. Several other Divisions change by only a few problems. Such small count differences determine the observed ordering but should not be read as independently established improvements. Section~\ref{sec:eval-mechanisms} examines how gains and losses can coexist even within a Division.

\begin{table}[htbp]
\caption{Results on the test sets: the complete official selections minus training problems. Official columns count official cvc5 \texttt{sat} answers within 24 seconds; tuned columns count locally validated answers within 24 seconds. The tuned configurations were evaluated during the last two calendar days of tuning.}
\label{tab:test}
\small\renewcommand{\arraystretch}{1.14}\setlength{\tabcolsep}{2.6pt}
\begin{tabularx}{\linewidth}{@{}Yrrrrrrrr@{}}
\toprule
& \multicolumn{4}{c}{2026 test} & \multicolumn{4}{c}{2025 test} \\
\cmidrule(lr){2-5}\cmidrule(lr){6-9}
Division & $n$ & Official & Tuned & $\Delta$ & $n$ & Official & Tuned & $\Delta$ \\
\midrule
Nonlinear integer arithmetic & 919 & 481 & 640 & +159 & 4,937 & 2,758 & 3,313 & +555 \\
ADT + bit-vectors & 931 & 824 & 879 & +55 & 4,486 & 3,989 & 4,308 & +319 \\
ADT + linear arithmetic & 517 & 501 & 505 & +4 & 535 & 535 & 535 & +0 \\
Bit-vectors & 1,230 & 1,140 & 1,142 & +2 & 7,359 & 6,892 & 6,808 & -84 \\
Datatypes & 537 & 310 & 314 & +4 & 1,715 & 1,128 & 1,137 & +9 \\
Equality & 500 & 500 & 500 & +0 & 1,461 & 1,461 & 1,461 & +0 \\
Equality + bit-vectors & 251 & 133 & 226 & +93 & 201 & 111 & 190 & +79 \\
Equality + linear arithmetic & 546 & 541 & 541 & +0 & 635 & 629 & 630 & +1 \\
Equality + nonlinear arithmetic & 272 & 200 & 203 & +3 & 232 & 201 & 201 & +0 \\
Floating point & 4,708 & 4,698 & 4,700 & +2 & 23,319 & 23,283 & 23,293 & +10 \\
Linear integer arithmetic & 1,004 & 754 & 826 & +72 & 3,857 & 3,124 & 3,477 & +353 \\
Linear real arithmetic & 354 & 303 & 324 & +21 & 419 & 355 & 387 & +32 \\
Nonlinear real arithmetic & 729 & 556 & 504 & -52 & 2,515 & 1,883 & 1,654 & -229 \\
\midrule
Diagnostic pooled total & 12,498 & 10,941 & 11,304 & +363 & 51,671 & 46,349 & 47,394 & +1,045 \\
\bottomrule
\end{tabularx}
\end{table}

\paragraph{Complete selections.}
On the complete selections, the tuned configurations validate 16,583 of 19,094 problems in 2026 and 52,965 of 59,750 in 2025, against 15,231 and 49,195 for official cvc5. These are 1,352 and 3,770 additional problems, or increases of 8.9\% and 7.7\% in the pooled total. Table~\ref{tab:full} shows that NIA, equality with bit-vectors, and LIA contribute large gains in both years. Because these sets include training problems, the larger improvements measure specialization to the complete workloads; the test-set results isolate performance beyond the training problems.

\begin{table}[htbp]
\caption{Results on the complete selections, which include the training problems. $O$ is the official cvc5 count and $T$ the tuned count for the same year and selection; $\Delta=T-O$ and change $=100(T-O)/O$.}
\label{tab:full}
\fontsize{8.5}{10}\selectfont\renewcommand{\arraystretch}{1.16}\setlength{\tabcolsep}{1.6pt}
\begin{tabularx}{\linewidth}{@{}Yrrrrrrrrrr@{}}
\toprule
& \multicolumn{5}{c}{2026 complete selection} & \multicolumn{5}{c}{2025 complete selection} \\
\cmidrule(lr){2-6}\cmidrule(lr){7-11}
Division & $n$ & Official & Tuned & $\Delta$ & Change & $n$ & Official & Tuned & $\Delta$ & Change \\
\midrule
Nonlinear integer arithmetic & 1,901 & 761 & 1,322 & +561 & +73.7\% & 7,516 & 2,940 & 5,034 & +2,094 & +71.2\% \\
ADT + bit-vectors & 1,513 & 1,201 & 1,365 & +164 & +13.7\% & 5,249 & 4,220 & 4,874 & +654 & +15.5\% \\
ADT + linear arithmetic & 798 & 736 & 758 & +22 & +3.0\% & 772 & 730 & 749 & +19 & +2.6\% \\
Bit-vectors & 1,909 & 1,674 & 1,712 & +38 & +2.3\% & 8,211 & 7,271 & 7,301 & +30 & +0.4\% \\
Datatypes & 871 & 461 & 497 & +36 & +7.8\% & 1,943 & 1,221 & 1,267 & +46 & +3.8\% \\
Equality & 714 & 714 & 714 & +0 & +0.0\% & 1,571 & 1,571 & 1,571 & +0 & +0.0\% \\
Equality + bit-vectors & 437 & 172 & 368 & +196 & +114.0\% & 475 & 132 & 326 & +194 & +147.0\% \\
Equality + linear arithmetic & 891 & 778 & 826 & +48 & +6.2\% & 891 & 778 & 831 & +53 & +6.8\% \\
Equality + nonlinear arithmetic & 506 & 282 & 319 & +37 & +13.1\% & 475 & 271 & 302 & +31 & +11.4\% \\
Floating point & 6,249 & 6,220 & 6,234 & +14 & +0.2\% & 24,369 & 24,293 & 24,332 & +39 & +0.2\% \\
Linear integer arithmetic & 1,770 & 1,091 & 1,329 & +238 & +21.8\% & 4,895 & 3,320 & 4,075 & +755 & +22.7\% \\
Linear real arithmetic & 599 & 453 & 519 & +66 & +14.6\% & 606 & 448 & 518 & +70 & +15.6\% \\
Nonlinear real arithmetic & 936 & 688 & 620 & -68 & -9.9\% & 2,777 & 2,000 & 1,785 & -215 & -10.8\% \\
\midrule
Diagnostic pooled total & 19,094 & 15,231 & 16,583 & +1,352 & +8.9\% & 59,750 & 49,195 & 52,965 & +3,770 & +7.7\% \\
\bottomrule
\end{tabularx}
\end{table}

\paragraph{Comparison with all participants.}
On the complete selections, which include the training problems, the tuned count exceeds every individual participant's 24-second count in two Divisions in both years. In floating point, it reaches 6,234 versus a maximum of 6,220 in 2026 and 24,332 versus 24,321 in 2025; in equality with nonlinear arithmetic, it reaches 319 versus 282 and 302 versus 271. These maxima include official cvc5; the best other participant reaches 6,207 and 24,321 in floating point and 282 and 209 in equality with nonlinear arithmetic. In 2026, the tuned count also exceeds every participant in ADT with linear arithmetic (758 versus 753) and ties the maximum in equality (714).

\paragraph{Output correctness and validator limits.}
Across both complete selections, 78,844 problems in total, the tuned configurations produce zero observed wrong answers and zero observed wrong models. Unverified models are not counted as correct. Table~\ref{tab:correctness} shows the 997 models excluded from validated counts. Most lack partial-function values; the rest cannot be parsed. Official cvc5 has the same validation error on 968 of these problems, so most such failures also occur without the tuned options. A Division with any such output would remain ineligible under the eligibility rule.

\begin{table}[htbp]
\caption{Output checks on the complete selections. Unverified models are excluded from every validated count.}
\label{tab:correctness}
\tight
\begin{tabularx}{\linewidth}{@{}Yrr@{}}
\toprule
Check & 2026 & 2025 \\
\midrule
Problems & 19,094 & 59,750 \\
Wrong answers or wrong models found & 0 & 0 \\
Missing partial-function values & 348 & 570 \\
Model parsing errors & 12 & 67 \\
Total unverified & 360 & 637 \\
Same validation error in official cvc5 & 342 & 626 \\
\bottomrule
\end{tabularx}
\end{table}

\subsection{Training Progress and Sustained Search}
\label{sec:eval-progress}
\begin{figure}[!htbp]
\centering
\includegraphics[width=\linewidth]{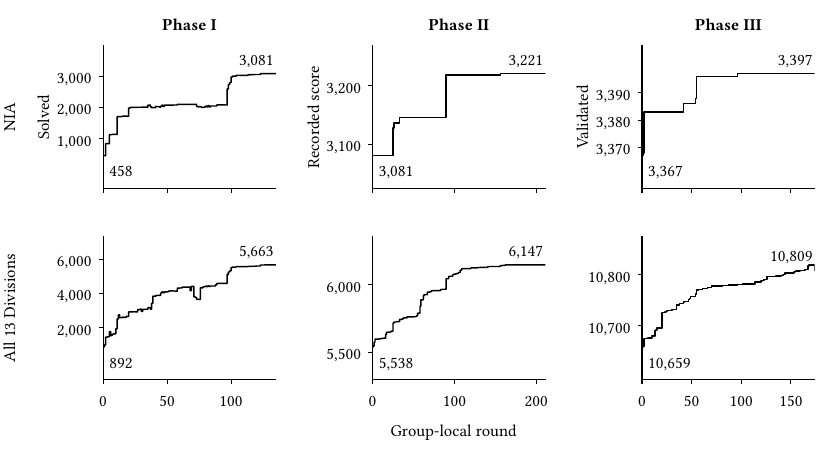}
\Description{Two rows and three phase columns show NIA and all thirteen Divisions including NIA. Each curve carries recorded incumbent scores across group-local rounds. Phase II begins at the recorded inherited scores.}
\caption{Reconstructed training scores for NIA (top) and all 13 Divisions including NIA (bottom). At $k$, the aggregate sums each Division's retained score after the $k$th group-local round of its group. Scores carry forward between supported updates and after a group ends; all measurements of a round share its position, while interrupted records and operator fixes occupy their own positions. Phase I counts solved problems for retained configurations, excluding crashes; documented infrastructure-failure runs carry the preceding score. Acceptances made while the phase-I score counted crashes are preserved, with their later fix. Phase II starts at the recorded inherited incumbent scores and uses validated counts for later retained updates. Phase III starts from remeasurement on the expanded workload and new hosts. These round-aligned sums of separate Division records are neither simultaneous whole-configuration evaluations nor the accounting-snapshot totals in the text. Within each column, the aggregate covers the NIA round range; later rounds are omitted. Vertical axes are independent; decreases and correctness fixes are retained.}
\label{fig:training}
\end{figure}

We trace how the tuned configurations evolved during tuning and what continued search contributed. Figure~\ref{fig:training} shows training progress across the three phases, with NIA as an example Division in the top row and the aggregate across all thirteen Divisions in the bottom row.

Each panel plots recorded incumbent scores against the group-local round. The aggregate aligns groups by their group-local rounds, sums their retained Division scores, and carries scores forward between updates. Within each phase, both rows show the same round range; later aggregate updates fall outside the plotted range. The phases use different scores and conditions: phase I counts solved problems, phase II begins with inherited scores and then uses validated counts, and phase III starts from remeasurement on the expanded workload and new hosts. The panels therefore describe progress within each phase; scores are not directly comparable across phases.

\paragraph{Progress through discrete improvements.}
The NIA curves show that improvement often arrives in steps, separated by stretches without a higher retained score. In phase I, an extended period near 2,000 solved problems is followed by a sharp rise above 3,000. Phase II also has a large jump after a plateau. Several rounds of search can therefore precede a configuration that materially improves the recorded result. The aggregate combines updates from different Divisions, so it rises more gradually, but it still shows some large jumps. Decreases in phase I need a separate explanation: the reconstruction excludes documented infrastructure-failure measurements but keeps acceptances made while the phase-I score counted crashes, together with their later fix.

\paragraph{Continued gains after earlier tuning.}
Progress continues in phase III, after the first two phases of tuning. Within the plotted range, NIA rises from 3,367 to 3,397 validated problems, and the aggregate rises from 10,659 to 10,809. The full records also contain accepted improvements beyond the plotted range: the bit-vectors Division improves at group-local round 181, and floating point records further improvements as late as round 547. A plateau in one Division does not mean that further search has exhausted improvements across the workload.

Late improvements become less frequent, but they are not uniformly small. In the accounting snapshot from the day before tuning ended, which covers 2,865 phase-III proposals, the acceptance rate outside floating point falls from 9\% in the first fifth of each group's proposals to about 1\% in each of the last two fifths. Of 57 Division acceptances, 46 initially improve the count by at most three problems, including 31 by exactly one. Yet bit-vector learned rewriting still yields a recorded gain of 30 validated problems.

Small recorded gains need careful interpretation. Repeated runs of identical floating-point configurations differ by up to four validated problems. Of the 57 acceptances, only one passes an uncorrected paired test, and none passes Holm correction. Later measurements that contain an accepted change exceed its original parent in 25 of 56 cases; these configurations may contain other edits, so the comparison is not an isolated retest of each change. Recorded gains support continued search, but they do not show that an individual gain is reproducible.

\paragraph{Validation failures during training.}
Across phases II and III, the correctness audit finds 6,661 wrong-model problem runs in 111 measurement executions. No Division with a wrong model was accepted; in 20 candidate rounds with a wrong model, a valid change in another Division was still retained. Validation and per-Division selection determined which changes survived. Appendix~\ref{app:training-validation} in the supplementary material details the option combinations associated with suspected cvc5 model-generation bugs.

\subsection{Interacting Algorithmic Choices}
\label{sec:eval-mechanisms}
Option changes can have large effects even though the solver implementation is unchanged. We investigate why through source analysis, controlled experiments, and family-level measurements. These studies show that configuration affects both access to existing strategies and the balance between benchmark families.

Theory and search options produced large gains in individual tuning rounds. On QF\_NIA, always using assertion relevance in nonlinear reasoning (\flag{nl-rlv=always}) gained 520 solved problems, and splitting variables on whether they are zero, trying zero first (\flag{nl-ext-split-zero}), gained 384. On QF\_ABV, disabling eager array-index splitting (\flag{no-arrays-eager-index}) gained 234 solved problems, and using justification only for early stopping (\flag{decision=stoponly}) gained 134. Each was a single-option change, and all four settings remain in the tuned configurations. Each gain is relative to the parent configuration of its tuning round, so the gains are not additive contributions to the test-set improvement.

\paragraph{Complementary strategies in the resulting configuration.}
The NIA configuration runs five portfolio workers, which combine arithmetic strategies with strategies that translate integer arithmetic to bit-vectors. Recorded tuning evaluations show a tradeoff in reducing the worker count. In two comparisons, four workers gained 32--37 ITS problems but lost 30--32 AProVE problems and nine leipzig problems, for net changes of $-2$ and $-13$. Six workers lost 144 validated problems in another comparison. These local observations show dependence on scheduling and benchmark family; they do not assign independent contributions to the final test gain.

\begin{figure}[!htbp]
\centering
\includegraphics[width=\linewidth]{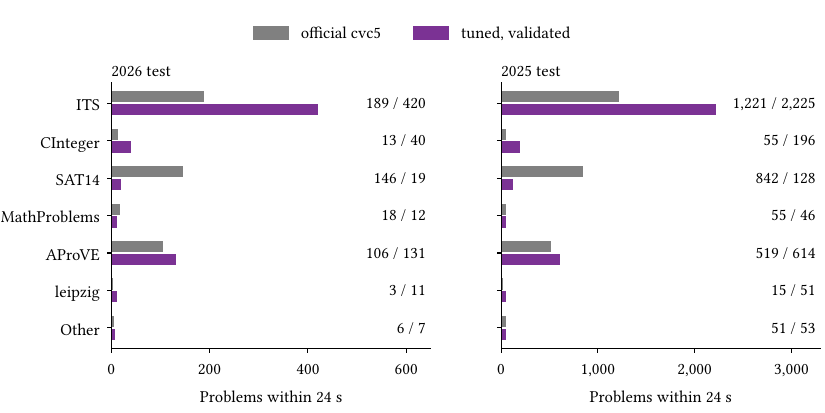}
\Description{NIA family-level counts on the 2026 and 2025 test sets, as two panels of grouped bars: each family has a grey bar for official cvc5 and a purple bar for the tuned configuration. ITS and CInteger gains coexist with large SAT14 losses in both years.}
\caption{NIA family tradeoffs in the final evaluations: 481 to 640 on 919 test problems in 2026 (+33.1\%), and 2,758 to 3,313 on 4,937 in 2025 (+20.1\%). In each pair the first value is the official cvc5 count and the second the tuned count. Other combines the remaining families. Values are recomputed from the per-problem records. Counts do not isolate the effect of any one option.}
\label{fig:nia-families}
\end{figure}

\paragraph{Family-dependent specialization.}
Figure~\ref{fig:nia-families} shows the tradeoff in the final test results. ITS improves from 189 to 420 in 2026 and from 1,221 to 2,225 in 2025, whereas SAT14 falls from 146 to 19 and from 842 to 128. In training rounds r000013, r000043, and r000153 of phase III for NIA, removing interval constraint propagation (\flag{nl-icp}) raised SAT14 from 21 or 22 to 105 or 106 of 110 problems but lost 88--99 ITS problems. SAT14 is about 2\% of the NIA training problems but 17\% of each year's NIA test set. Interval constraint propagation is therefore a plausible contributor to the test regression, although the observations do not establish this attribution without a test-set intervention.

\paragraph{Controlled experiments.}
For NIA and LIA, we compare configurations on twenty problems per logic selected from recorded tuned successes. Variants run back-to-back on the same assigned cores. The reference portfolio is cvc5's portfolio with the initial configuration's options for the logic. The interventions change scheduling or remove a model-output incompatibility and keep the reference portfolio's other relevant behavior. The internal time-limit intervention is explanatory, and it is not an allowed tuning option. Source analysis and these experiments show the mechanisms; they do not decompose every final performance gain. Table~\ref{tab:mechanisms} reports \texttt{sat} answers within 24 seconds for these selected problems. The experiment asks whether the suspected barriers can prevent useful strategies from running; it does not estimate their frequency across the complete workload.

\begin{table}[htbp]
\caption{Configuration determines access to strategies and usable outputs. The rows come from controlled experiments on 20 problems per logic, selected from recorded tuned successes; values are \texttt{sat} answers within 24 seconds, not generalization rates. Variants ran back-to-back on loaded hosts. The portfolio changes examined here were discovered by the agents during tuning; these post-hoc experiments investigate why those changes helped and are not agent probes during tuning.}
\label{tab:mechanisms}
\tight
\begin{tabularx}{\linewidth}{@{}P{.12\linewidth}YP{.16\linewidth}P{.22\linewidth}@{}}
\toprule
Case & Configuration / mechanism & \texttt{sat} within 24 s & Additional observation \\
\midrule
NIA & Reference portfolio: internal limit 1,200 s & 1 / 20 & 0 / 20 started a second strategy. \\
& Reference + experimental internal 24 s limit & 8 / 20 & 20 / 20 started multiple strategies. \\
& Reference + five portfolio workers & 12 / 20 & 20 / 20 started multiple strategies. \\
& Tuned configuration & 20 / 20 & Answers came from five strategies. \\
\midrule
LIA & Reference portfolio: model-output option conflict & 5 / 20 & Fallback occurred on all 20. \\
& Reference + experimental internal 24 s limit & 5 / 20 & Time limit does not fix the conflict. \\
& First strategy without conflicting option; portfolio off & 9 / 20 & Exposes the intended strategy. \\
& Tuned configuration & 12 / 20 & All 20 solved in the recorded selection run. \\
\bottomrule
\end{tabularx}
\end{table}

\paragraph{Making solver capabilities accessible.}
The controlled experiments identify scheduling and compatibility constraints (Table~\ref{tab:mechanisms}). NIA's reference internal schedule can allocate 420 seconds to its first strategy, although the external limit is 24 seconds. A reference LIA strategy contains an option incompatible with model generation, which causes fallback. For linear real arithmetic (LRA), source analysis shows that a forked worker writes its model to a pipe and can block on a large model before the parent process reads the pipe. On the 2026 test set, 14 of the 21 QF\_LRA problems validated only by the tuned configuration produced output logs above 64~KiB, versus none of the 243 validated by both. Official cvc5 reached its 1,200-second timeout on all 14, while the tuned configuration answered them in a median of 12 seconds. QF\_UFBV shows the same pattern: 79 of the 94 problems validated only by the tuned configuration produced output logs above 64~KiB, versus none of the 132 validated by both. Official cvc5 reached its 1,200-second timeout on all 79, while the tuned configuration, which disables the QF\_UFBV portfolio, answered them in a median of 0.9 seconds. Log sizes include a small BenchExec header. The timing contrast and the output-size pattern strongly support the pipe-blocking explanation, although builds and hosts differ. The selected 20-problem experiments explain access to solver capabilities; they do not estimate effects on the complete workloads or reproduce the tuning campaign.

In the NIA experiment, an internal 24-second limit lets the reference portfolio start several strategies on every selected problem and raises the answer count from one to eight. Running five workers yields twelve. In LIA, changing the internal limit leaves the count at five, but removing the incompatible option and running the intended strategy directly raises it to nine. The different responses separate a scheduling barrier in NIA from an option-compatibility barrier in LIA.

The agents discovered both portfolio changes during tuning. Switching the portfolio off gained 327 solved problems on QF\_LIA, 161 on QF\_UFBV, and 125 on QF\_ABV, each in a single-option round, and 26 validated problems on QF\_LRA. On QF\_NIA, changing from sequential execution to four parallel portfolio workers gained 255 solved problems; a later increase from four to five gained 61. These are observed gains in the respective tuning rounds, not a decomposition of the final test gain. Together with the family tradeoffs, these changes explain why tuning can improve an existing solver without introducing a new decision procedure: it changes which capabilities become available under the target workload and resource limits.

\subsection{Hypothesis-Based Proposals}
\label{sec:eval-hypotheses}
The traces also show whether agents use accumulated evidence to formulate hypotheses about configuration changes. We analyze recorded text from all three phases and extract each hypothesis about a configuration effect with the evidence it cites. The extracted hypotheses include prospective hypotheses, stated before the associated measurement, and retrospective explanations; quotation timestamps distinguish the two. Primary annotations were produced with DeepSeek-V4.1-Flash-1017. As a second reader, Claude annotated phase III through its CLI; this invocation did not pin an exact model version. This analysis concerns what the agents record during search; it does not infer an unrecorded reasoning process.

\paragraph{Coverage and evidence.}
Table~\ref{tab:hyp-coverage} separates the availability of agent text from the presence of an extracted hypothesis. Text is available for 4,489 of 4,559 records (98.5\%); the other 70 have none. A prediction is present in 752 records, absent in 3,767, and unclear in 40 because of missing evidence. Across the campaign, the extraction identifies 964 hypotheses in all six tuning groups, of which 901 were stated before their measurement started; the remaining 63 are not prospective. Hypothesis-based proposals are therefore a recurring component of the search, although not every proposal contains an explicit prediction.

\begin{table}[htbp]
\caption{Hypothesis extraction over all 4,559 records. Text coverage and prediction presence are separate labels; timing is assessed per hypothesis.}
\label{tab:hyp-coverage}
\tight
\begin{tabularx}{\linewidth}{@{}Yrrrrr@{}}
\toprule
Phase & Records & Text present & With predictions & Hypotheses & Prospective \\
\midrule
I & 517 & 502 & 95 & 101 & 63 \\
II & 997 & 952 & 104 & 127 & 102 \\
III & 3,045 & 3,035 & 553 & 736 & 736 \\
\midrule
Total & 4,559 & 4,489 & 752 & 964 & 901 \\
\bottomrule
\end{tabularx}
\end{table}

Prior measurements are the most common evidence category: 620 of the 964 extracted hypotheses (64.3\%) draw on earlier experimental results. Source or documentation inspection accounts for another 203 (21.1\%). The distribution matches the workflow design: persistent measurements give later sessions evidence, while source code and option definitions supply explanations that can motivate new changes. The proportions describe the extracted hypotheses, including those not marked as prospective, rather than all proposal rounds.

\paragraph{Hypotheses and measured outcomes.}
We compare the predicted direction of change with the candidate's recorded result relative to its parent configuration. The comparison uses solved counts in phase I and validated counts in later phases. The parent's result is the measurement of the retained configuration. No comparison is made without matching recorded parent options or a common count metric. A noise-aware labeling rule distinguishes supported, contradicted, mixed, and inconclusive outcomes; a hypothesis without an applicable comparison is not evaluated. Table~\ref{tab:hyp-outcomes} reports these categories.

A predicted increase or decrease is supported or contradicted at a signed change of at least $2s$; a predicted no-change is supported within $s$ and contradicted at or beyond $3s$. Here $s$ is the population standard deviation of count differences between repeated measurements of identical recorded Division options, with a floor of 1; it is not a single-run standard deviation. Four Divisions without sufficient pairs use $s=4$. Thresholds for a target logic use the corresponding differences from these same pairs. The rule is a heuristic, not a significance test. The automated labels depend on the extracted direction, target, and parent comparison.

\begin{table}[htbp]
\caption{Hypothesis-level outcome labels across all three phases, including 63 non-prospective hypotheses, all labeled not evaluated. The rule assesses the direction of parent-relative solved-count changes in phase I and validated-count changes in phases II--III; it does not independently establish a mechanism, family-specific prediction, or predicted effect size.}
\label{tab:hyp-outcomes}
\tight
\begin{tabular*}{\linewidth}{@{\extracolsep{\fill}}rrrrrr@{}}
\toprule
Hypotheses & Supported & Contradicted & Mixed & Inconclusive & Not evaluated \\
\midrule
964 & 96 & 113 & 4 & 258 & 493 \\
\bottomrule
\end{tabular*}
\end{table}

Of the 471 hypotheses assigned an outcome label, 96 are supported and 113 contradicted; four are mixed and 258 inconclusive. Another 493 are not evaluated. The traces therefore contain testable hypotheses, including hypotheses that measurement contradicts, rather than only explanations attached to successful edits. Because many hypotheses are inconclusive or not evaluated, the table is not a simple estimate of prediction accuracy.

\paragraph{Reliability and interpretation.}
Table~\ref{tab:hyp-reliability} compares the extraction with a second reader for phase III. The two readers agree on whether a record contains a prediction in 93.3\% of records, but their lists of predicted directions agree exactly in 51.7\% of the records that both flag. Detecting a prediction is therefore more stable than assigning its direction and target. We use the analysis to show the presence and evidence sources of hypothesis-based proposals; it does not establish that these proposals outperform alternatives or that an observed count change confirms the proposed mechanism.

No separate measurement-result fields were supplied to the annotator, but some recovered transcripts include subsequent results; extraction was therefore not uniformly outcome-blind. Even with exact quotations, not every target or direction interpretation is reliable. A check against raw measurement-start timestamps confirms that all 901 quotations marked prospective precede their measurements.

\begin{table}[htbp]
\caption{Evidence used in all 964 hypotheses, and a second-reader comparison restricted to phase III. Comparisons match the extraction with the available second-reader annotations by round ID.}
\label{tab:hyp-reliability}
\tight
\begin{tabularx}{\linewidth}{@{}P{.37\linewidth}P{.22\linewidth}Y@{}}
\toprule
Quantity & Value & Interpretation \\
\midrule
Prior measurements & 620 (64.3\%) & Most common evidence category. \\
Source or documentation inspection & 203 (21.1\%) & Solver behavior or option definitions motivate proposals. \\
Theoretical prior / other & 134 / 7 & One evidence label per hypothesis. \\
Agreement on prediction presence & 2,791 / 2,992 (93.3\%) & Both readers flag 402 records. \\
Direction-list agreement when both flag & 208 / 402 (51.7\%) & Sorted direction lists match exactly, counting repeated directions. \\
\bottomrule
\end{tabularx}
\end{table}

\subsection{Diagnostic Reasoning and Workflow Authority}
\label{sec:eval-diagnostics}
The traces also show agents recognizing limitations of their own search. Table~\ref{tab:authority} lists recorded diagnoses of measurement noise and family tradeoffs, alongside the workflow decisions that govern acceptance and continuation. These observations help distinguish an agent's ability to diagnose a problem from the workflow's ability to act on that diagnosis.

\begin{table}[!htbp]
\caption{Diagnosis and continued search within phase III. Recorded observations distinguish an agent's ability to recognize a limitation from a workflow decision to change the acceptance rule or stop searching.}
\label{tab:authority}
\tight
\begin{tabularx}{\linewidth}{@{}P{.20\linewidth}Y Y@{}}
\toprule
Observation & Recorded evidence & Workflow boundary \\
\midrule
NIA repeat-measurement diagnosis & In a session midway through phase III, the agent reported ten runs of configurations equivalent to the incumbent, with mean 3,386 and standard deviation about 5.4; the incumbent record of 3,396 was their maximum. & Acceptance still compared candidates with the incumbent record. Recognizing noise did not itself change that rule. \\
NIA family tradeoff & The same session described removing \flag{nl-icp} as solving SAT14 while losing ITS. Recorded training comparisons show the tradeoff. & The measured net score governed acceptance; no family-preservation rule was evaluated. \\
Floating-point continued search & The accounting snapshot from the day before tuning ended records 1,361 rounds and about 229 measurement hours after the last acceptance. & The workflow had no automatic stopping rule. This observation does not establish an optimal stopping threshold. \\
\bottomrule
\end{tabularx}
\end{table}

In NIA, an agent compared repeated measurements of the same effective configuration and recognized that the incumbent record exceeded the mean of the repeats. The same session described an option change that helped one family but hurt another. Both observations are relevant to future proposals, but neither changes the acceptance rule: the harness still compares counts against the incumbent record, and there is no separate requirement to preserve each family's score. In floating point, the search continued for 1,361 rounds and about 229 measurement hours after the last acceptance in the accounting snapshot.

These cases show that an agent's diagnosis can be useful while the search continues inefficiently. The workflow preserves evidence and constrains which edits agents can make, but it does not automatically turn a diagnosis into a new stopping rule or a different statistical criterion. The limit is structural. Agents reason within the rules they are given, and changing those rules remains a separate design decision.

\section{Related Work}
\label{sec:related}

\paragraph{\normalfont\bfseries Automated Solver Configuration and Algorithm Selection.}
Automated configuration search looks for parameter settings that perform well on a workload. ParamILS uses iterated local search~\cite{hutter2009paramils}, while SMAC uses models of configuration performance to guide candidate selection~\cite{hutter2011smac}. Tuning SPEAR showed that the approach works on verification workloads~\cite{hutter2007spear}. Our task has the same goal of specializing a solver to a workload. Agents propose command-line option edits within a bounded interface and choose candidates using documentation, source code, and prior measurements. The study examines how this proposal process operates within a sustained workflow, including the evidence agents use and the configurations selected by the harness.

Algorithm selection and strategy synthesis specialize at different levels. MachSMT ranks solvers for an individual problem by predicted runtime~\cite{scott2021machsmt}. FastSMT learns formula-transformation policies and synthesizes branching strategy programs that guide an existing solver~\cite{balunovic2018fastsmt}. Our configurations assign options by Division and logic, including controls over cvc5's existing portfolio. They apply the same effective options to every problem of a logic, without per-problem solver selection or synthesis of a new strategy program. This scope lets the study examine how existing solver mechanisms become effective under a short time budget and model-validation requirements.

\paragraph{\normalfont\bfseries LLM-Assisted Solver Improvement.}
LLM-based configuration also targets mixed-integer programming (MIP) solvers. GRIMIP uses an LLM as a probabilistic surrogate within Bayesian optimization for per-problem configuration~\cite{luo2026grimip}. In concurrent work, LDST combines structured SCIP log diagnostics, parameter meanings, and bounded memory of best and recent trials to propose sparse edits around the best configuration~\cite{you2026ldst}. Its experiments include small training sets of several problems and within-family transfer, as well as matched-budget comparisons with classical configurators. LDST brings evidence-guided configuration search to a different solver domain. Our study examines SMT specialization through source-informed investigation of option interactions, access to portfolio strategies, and benchmark-family tradeoffs. In our final workflow, the harness independently validates returned models and retains edits per Division.

Prior work on SAT solvers changes heuristic implementations. AutoSAT generates heuristic code within a modular search space, guided by hill climbing or evolutionary search~\cite{sun2024autosat}. AutoModSAT combines a deliberately modular solver with heuristic presearch, prompt optimization, and evolutionary code generation; it also checks satisfying assignments and UNSAT proofs~\cite{sun2026automodsat}. We specialize an existing mature cvc5 build through command-line options and leave its implementation unchanged, rather than restructuring it for generated heuristics. The resulting analysis concerns how agents make existing solver capabilities available and combine them during sustained tuning.

\paragraph{\normalfont\bfseries General LLM Optimization and Iterative Experimentation.}
Several LLM optimization methods rely on evaluation feedback. OPRO supplies earlier candidate solutions and their scores in the prompt to generate subsequent candidates~\cite{yang2024opro}. FunSearch combines LLM-based program generation with automated evaluation and an evolving program population~\cite{romeraparedes2024funsearch}. AlphaEvolve uses an evolutionary coding agent to improve algorithms through code changes and evaluator feedback~\cite{novikov2025alphaevolve}.

Database tuning also uses LLM knowledge in configuration search. GPTuner extracts guidance from manuals and discussions to select which parameters to tune and narrow their value ranges before Bayesian optimization~\cite{lao2024gptuner}. $\lambda$-Tune generates complete database configuration scripts from task context and evaluates a small candidate set~\cite{giannakouris2025lambdatune}.

Our workflow shares this propose-and-evaluate pattern and applies it to solver configurations. Fresh agent sessions can inspect persistent configurations, per-problem outcomes, and rejected proposals alongside solver source and documentation. The harness supplies controlled complete-group measurements and explicit state updates, which connect open-ended proposal generation to a consistent experimental procedure. The resulting history supports analysis of the hypotheses agents recorded and the evidence available to them during sustained experimentation.
\section{Conclusion}
We presented a workflow in which general-purpose LLM agents specialize a mature SMT solver through configuration tuning. The tuned cvc5 configurations improve performance across multiple Divisions and in both benchmark years, so an existing solver's configuration interface still offers useful gains. Our studies show how tuning can make existing algorithms effective under workload and resource constraints by changing access to strategies, scheduling, and option compatibility. The experiment traces also show a hypothesis-based component of the search: agents use accumulated measurements and solver knowledge to motivate later proposals. SMT solver configuration is therefore a productive setting for agent-driven experimentation: agents explore possible improvements, and measurements with independent model checks tie the search to observed results.
\paragraph{AI Use.}
This research studies the use of OpenAI Codex and Anthropic Claude agents for SMT solver configuration tuning; their roles in the experiments are described in Sections~\ref{sec:approach} and~\ref{sec:evaluation}. We also used DeepSeek-V4.1-Flash-1017 for primary trace annotation and Claude for a second-reader comparison, under human supervision, as described in Section~\ref{sec:eval-hypotheses}.

\section*{Data-Availability Statement}
Upon publication, we plan to publicly release the tuning framework implementation, the tuned configurations, and all experimental data and results supporting this paper. We also plan to submit these materials for artifact evaluation.

\bibliographystyle{ACM-Reference-Format}
\bibliography{refs}

\appendix
\clearpage
\noindent{\Large\bfseries Supplementary Material}\par\bigskip
\section{Configuration Scope}
\label{app:configuration-scope}
Table~\ref{tab:configuration-scope} organizes the configuration scope by solver mechanism and distinguishes the tunable options from those changed in the tuned configurations.

Tunable options are the command-line options that the target build defines and lists in its help output, excluding nine harness-owned options such as time limits and the input language. We count each distinct option once, treating aliases and the positive and negative forms of an option as one option, and assign it to one category. An option is changed if its effective value differs from the initial configuration for at least one logic. The Other category consists mainly of options for quantifiers, strings, proofs, and input and output.

\begin{table}[htbp]
\caption{Tunable options and options changed in the tuned configurations, by category. Counts are distinct options.}
\label{tab:configuration-scope}
\tight
\begin{tabularx}{\linewidth}{@{}P{.21\linewidth}YP{.14\linewidth}P{.16\linewidth}@{}}
\toprule
Category & Description & Tunable options & Changed options \\
\midrule
Random seeds & Seeds controlling randomized choices. & 2 & 1 \\
Portfolio & Use of the predefined portfolio and number of concurrent portfolio workers. & 2 & 2 \\
SAT back ends & Choice of SAT and bit-vector solvers; back-end simplification. & 4 & 3 \\
SAT search & Restarts, randomized search, and conflict or lemma processing. & 11 & 5 \\
Decision heuristics & Decision modes and justification heuristics. & 4 & 3 \\
Preprocessing & Assertion simplification, rewriting, and other preprocessing options. & 20 & 13 \\
Model output & Model representation, printing, and checking. & 8 & 0 \\
Nonlinear arithmetic & Nonlinear procedures and arithmetic encodings. & 27 & 12 \\
Other arithmetic & Arithmetic procedures, simplex, cuts, and approximation. & 37 & 22 \\
Bit-vectors & Bit-vector reasoning, bit-blasting, and encodings. & 14 & 8 \\
Arrays & Array-theory procedures. & 7 & 6 \\
Datatypes & Datatype-theory procedures. & 10 & 3 \\
Floating point & Floating-point procedures. & 2 & 0 \\
Other & Options not matched by the preceding categories. & 346 & 0 \\
\midrule
Total & Distinct options across all categories. & 494 & 78 \\
\bottomrule
\end{tabularx}
\end{table}

\clearpage
\Needspace{20\baselineskip}
\section{Training Validation and Suspected Solver Bugs}
\label{app:training-validation}
Across phases II and III, the recorded correctness audit finds 111 measurement executions containing 6,661 wrong-model problem runs. No Division with a wrong model was accepted. In 20 distinct candidate rounds with wrong models, the harness still accepted a different Division with correct models. Per-Division selection can therefore retain a valid change elsewhere without publishing the failing Division's edit.

Wrong models cluster around a few option combinations:
\begin{itemize}
\item \flag{--dt-binary-split} occurs in all observed wrong-model measurement/logic pairs for QF\_DT and QF\_UFDT in the phase-I--II archive, including retrospective validation, and in seven of 15 such pairs in phase III.
\item \flag{--theoryof-mode=term} is associated with wrong models on QF\_ABV and QF\_UFBV.
\item \flag{--ee-mode=central} is associated with wrong models on QF\_UFDTLIA.
\item \flag{--deep-restart=all} is associated with a wrong model on QF\_UFNIA.
\end{itemize}
These are suspected model-generation bugs in the target build; association alone does not identify a minimal cause. Model validation makes such failures visible during configuration search, and the eligibility rule prevents them from being accepted.

\end{document}